\documentclass{article}

\PassOptionsToPackage{numbers,sort&compress}{natbib}
\usepackage{arxiv}

\usepackage[utf8]{inputenc} 
\usepackage[numbers,sort&compress]{natbib}
\usepackage{xurl}           
\usepackage[T1]{fontenc}    
\usepackage{hyperref}       
\usepackage{url}            
\usepackage{booktabs}       
\usepackage{tabularx}       
\usepackage{amsfonts}       
\usepackage{nicefrac}       
\usepackage{microtype}      
\usepackage{lipsum}
\usepackage{graphicx}
\graphicspath{ {./images/} }

\title{Language Predicts Identity Fusion Across Cultures and Reveals Divergent Pathways to Violence}

\author{%
Devin R. Wright$^{1,2,3\ast}$\And
Justin E. Lane$^{3,4,5}$\And
F. LeRon Shults$^{3,4,6,7}$ \And
\\$^{1}$Center for Complex Networks and Systems Research, Luddy School of Informatics, Computing, and Engineering, \\
Indiana University, Bloomington \& 47405, USA.\\
$^{2}$Cognitive Science Program, Indiana University, Bloomington \& 47405, USA.\\
$^{3}$CulturePulse, Bratislava \& 821 01, Slovakia.\\
$^{4}$Scientific Advisory Board, DEKK Institute, Bratislava \& 811 05, Slovakia.\\
$^{5}$Institute of Ethology and Social Anthropology, Slovak Academy of Sciences, Bratislava \& 814 38, Slovakia.\\
$^{6}$Department of Global Development and Social Planning, University of Agder, Kristiansand \& 4605, Norway.\\
$^{7}$Health and Social Sciences, NORCE Center for Modeling Social Systems, Bergen \& 5838, Norway. \\
Corresponding author. Email: \texttt{devrwrig@iu.edu}
}

\begin{document} 

\maketitle

\begin{abstract} \bfseries \boldmath

In light of increasing polarization and political violence, understanding the psychological roots of extremism is increasingly important. Prior research shows that \textit{identity fusion} predicts willingness to engage in extreme acts. We evaluate the Cognitive Linguistic Identity Fusion Score, a method that uses cognitive linguistic patterns, LLMs, and implicit metaphor to measure fusion from language. Across datasets from the United Kingdom and Singapore, this approach outperforms existing methods in predicting validated fusion scores. Applied to extremist manifestos, two distinct high-fusion pathways to violence emerge: ideologues tend to frame themselves in terms of group, forming kinship bonds; whereas grievance-driven individuals frame the group in terms of their personal identity. These results refine theories of identity fusion and provide a scalable tool aiding fusion research and extremism detection.

\end{abstract}

\noindent

\noindent

In light of recent high profile assassinations, assassination attempts, and political violence linked to a variety of digital platforms from online gaming to social media use, understanding the risk of real world violence that results from radicalization online has never felt more important. In the offline world, cognitive scientists have focused on a particularly powerful psychological predictor of endorsement of violent responses called Identity Fusion~\cite{SwannGomezEtAl2009,SwannJettenGomezWhitehouseBastian2012,SwannKleinGomez2024,GomezBrooksEtAl2011}, which is a form of extreme alignment with a group whereby individuals are unable to effectively separate themselves from their fusion target (e.g., a social group, group-defining beliefs, values, an ideology, leader, or other abstract target). This framework has been shown to be a powerful explanation for both state and non-state violent group actors as well as lone-wolf actors~\cite{SwannKleinGomez2024,Kristinsdottir_Ebner_Whitehouse_2025, EbnerKavanaghWhitehouse2022}. This article demonstrates the superior analytic and predictive capacity of a new
approach~\cite{wright2025clifs} to profiling the psychological signatures common among violent extremists in ways that are more cognitively informed than past approaches.

Previous attempts at preventing online extremism have generally been reliant on content moderation~\cite{Goldstein2023UnderstandingT,laneetal2021}, which has failed to date on several dimensions. First, the volume of social media posts on even moderately sized platforms is too much for human centric content moderation. 
For example, Facebook has an estimated $\approx 5m$ discrete interactions (posts, comments, likes, messages, etc.) every minute, which would require an inordinate number of human content moderators to cover them all~\cite{Hooda_2025}. Second, systems trained to classify hate speech and other violations of common terms of service (such as sharing violent manifestoes), are overly reliant on past data. Even if such approaches were effective in stopping potential copy-cat incidents and similar radical profiles, they are not able to spot extremists with different profiles who may be violently respondent to new issues. By addressing the psychological mechanisms that are known to predict violent extremism, as opposed to political or other cultural topics that may be tied to an extreme position, new AI systems could be deployed to better address these shortcomings and triage the massive amount of controversial data to focus on only those users who show signatures of high identity fusion.

It is well known across the AI industry that one of the main practical challenges to
creating a scalable system is ``model overfit.'' Models may fit \textit{training} and \textit{test} data extremely well, but perform far less accurately on other data in production environments. To address this challenge, we tested a novel approach utilizing out-of-domain data from earlier studies where psychometrically validated measurements of identity fusion were collected in conjunction with interviews about participants' religious beliefs.

\subsection*{Data}

The first dataset was drawn from prior unpublished work regarding identity fusion and significant religious experience and was collected via Amazon Mechanical Turk (MTurk). After excluding participants who did not complete the survey ($N = 680$) or report a significant religious experience ($N = 256$), the final sample comprised 521 individuals (218 male, 303 female), with 283 residing in Singapore and 238 in the UK. Participants ranged in age from 18--84 years ($M = 24.18$, $SD = 13.27$) and represented multiple denominations, Roman Catholic ($20.9\%$), Anglican ($18\%$), independent/non-denominational ($10.9\%$), Methodist ($8\%$), Baptist ($5.4\%$), and no other denomination or affiliation ($5\%$) in the sample. \textit{For the current study}, we focus on analyzing participants who shared written descriptions of either ``conversion'' or ``reaffirmation'' experiences, as well as corresponding  Verbal Identity Fusion Scale (VIFS; $N = 85$) scores~\cite{GomezBrooksEtAl2011}.

For the second dataset, data were collected during eight months of fieldwork in Singapore from participants recruited at two churches. Participants completed an English-language semi-structured interview and survey. Interviews elicited interpretations of Sunday Services, a shared low-arousal weekly Christian ritual, by asking participants to describe its purpose and the meaning of its liturgical elements; behavioral descriptions were excluded from coding. After removing one outlier and incomplete cases, the final sample comprised $N = 33$ participants (12 female, 21 male; age $M = 25.87$, $SD = 3.84$). All participants were fluent in English; twenty-two spoke English at home, two Mandarin, one Indonesian, and the rest a mix of English and a Chinese dialect (``Singlish''). The survey included the Postmes four-item social identification scale~\cite{postmes2013additional,doosje1995perceived}, the VIFS~\cite{GomezBrooksEtAl2011}, ritual attendance frequency, demographics, and measures of socio-economic status, political and theological beliefs, and personal religiosity.

The third dataset builds on the extremist manifesto corpus introduced in prior work, which contains fifteen manifestos \cite{EbnerKavanaghWhitehouse2022,EbnerKavanaghWhitehouse2024b}. For the present study, we retain only the nine entries labeled as ``Violent Self-Sacrificial'' or ``Ideologically Extreme.'' We further augment this set with nine additional violent manifestos collected by CulturePulse, including four cases in which individuals are primarily motivated by perceived victimhood. See Table~\ref{tab:manifestos} for manifestos included~\cite{methods}.

\subsection*{Method: CLIFS}

The Cognitive Linguistic Identity Fusion Score (CLIFS) estimates an individual's level of identity fusion using written text alone. CLIFS previously outperformed prior methods and human annotation in predicting identity fusion on in-domain data and improved violence-risk prediction by more than $240\%$~\cite{wright2025clifs}. The model integrates several complementary components: (i) cognitive linguistic patterns identified in prior research; (ii) opaque embedding features from large language models to capture signals not yet described in the literature; (iii) coarse-grained fusion-level probabilities from a fine-tuned ModernBERT classifier; and (iv) masked language model (masked-LM) based implicit metaphor detection~\cite{GomezBrooksEtAl2011,AshokkumarPennebaker2022,CardChangEtAl2022,EbnerKavanaghWhitehouse2022,EbnerKavanaghWhitehouse2022c,EbnerKavanaghWhitehouse2024,EbnerKavanaghWhitehouse2024b,modernbert,wright2025clifs}.

A central innovation of CLIFS is its use of implicit metaphor detection to capture how individuals conceptualize their fusion target in relation to the self---a method also previously employed to detect dehumanizing speech in political discourse~\cite{CardChangEtAl2022,wright2025clifs}. Building on this approach, CLIFS uses implicit metaphor detection to derive three key measures: \textit{Fictive-Kinship} ($K_f$; the extent to which individuals view their group as family), \textit{Fusion Proximity} ($f_{(I,T)}$; a bidirectional measure of self-group conceptual overlap), and directional identity scores ($S_{I \to T}$ and $S_{T \to I}$), indexing perceived identity overlap from self to group and group to self~\cite{wright2025clifs,CardChangEtAl2022}. In ablation studies, these features produced the largest performance gains among all non-trained feature types~\cite{wright2025clifs}. Remarkably, the four implicit metaphor features outperformed high-dimensional semantic embeddings (768 dimensions), highlighting the effectiveness of metaphor-based signals for detecting identity fusion~\cite{wright2025clifs}.

For complete details of the CLIFS methodology---including vocabulary construction, ablations, data and augmentation strategies, in-domain comparisons with prior methods and human coding, and downstream violence-risk prediction---see the following~\cite{methods, wright2025clifs, CLIFS_github}.
 
\subsection*{Experiments}

\subsubsection*{Cross-Cultural Out-of-Domain Validation}

To validate that CLIFS captures identity fusion itself---rather than solely fitting held-out test data---and to further mitigate potential overfitting, we evaluate it on the two out-of-domain datasets described above. These datasets span cultural contexts, countries, time periods (data from over a decade prior to the current study), and communication domains not represented in the original CLIFS study~\cite{wright2025clifs}.

A challenge arises with the first dataset (UK \& Singapore MTurk data, $N = 85$): while many responses are rich, some entries are too short or lack sufficient substance for CLIFS to analyze (e.g., ``i like jesus so i am converted,'' ``My pray has come true,'' or simply ``no''/``nice''). Because CLIFS relies on semantic pattern analysis rather than simple word-matching, these minimal responses cannot be meaningfully processed. To address this, we filter the data at both the word and sentence level using the \textsc{Natural Language Toolkit} (\textsc{NLTK})~\cite{Bird_Natural_Language_Processing_2009}. Specifically, we conduct three evaluations by applying word-count thresholds (minimums of 30, 40, and 50 words; $N = 50$, $40$, and $25$) and three evaluations by applying sentence-count thresholds (minimums of 2, 3, and 4 sentences; $N = 58$, $37$, and $21$).  

We perform two evaluations using the second dataset (Singapore field data). First, each transcript is analyzed as a single entry ($N = 33$). Second, because many transcripts exceed the 384-token input limit of the Sentence Transformer model used for CLIFS opaque embeddings (\texttt{all-mpnet-base-v2}), we apply a chunking strategy to prevent content loss~\cite{reimers-2019-sentence-bert,song2020mpnet,wright2025clifs}. Texts are segmented into sentence-preserving chunks of up to 300 words using \textsc{NLTK}~\cite{Bird_Natural_Language_Processing_2009}. Each chunk is treated as a separate entry, yielding $N = 115$ samples, which are filtered to retain only entries with at least four sentences (final $N = 109$).  

For all entries, we obtain CLIFS predictions as well as baseline identity fusion metrics for comparison. These baselines include the Unquestioning Affiliation Index (UAI), its simplified variant without $z$-scores (nUAI), and the fusion submodule of the Violence Risk Index (VRI-Fusion)~\cite{AshokkumarPennebaker2022,EbnerKavanaghWhitehouse2024,EbnerKavanaghWhitehouse2024b,EbnerKavanaghWhitehouse2022,EbnerKavanaghWhitehouse2022c,wright2025clifs}. The UAI measures identity strength using the \textsc{Linguistic Inquiry and Word Count} (\textsc{LIWC}) software and was tuned against VIFS scores, and the nUAI was introduced as a baseline in the original CLIFS work~\cite{pennebaker2015development,AshokkumarPennebaker2022,wright2025clifs}. VRI-Fusion is a dictionary-based metric that counts the proportion of sentences containing kinship words manually identified as fusion markers in prior research~\cite{EbnerKavanaghWhitehouse2024,EbnerKavanaghWhitehouse2024b,EbnerKavanaghWhitehouse2022,EbnerKavanaghWhitehouse2022c}.

\subsubsection*{Pathways to Violence}

We use the extremist manifesto dataset to examine two presumed pathways to extreme violence: one associated with identity fusion and one not. While most cases are ideologically motivated, a subset links violence to perceived personal victimization (e.g., bullying, wrongful termination, incel), which we group into a single ``Victim'' class.

Our hypothesis is that CLIFS scores will be lower for individuals in the Victim class compared to those in the ideologically driven class (``Ideologue''). This expectation reflects the absence of a broader group identity or ideology for victimhood-driven actors to fuse with. To test this, we apply the same chunking procedure as before, limiting samples to a maximum of 300 words while preserving full sentences ($N =5,263$).

\subsection*{Results}

\subsubsection*{Cross-Cultural Generalization}

\begin{figure} 
	\centering
	\includegraphics[width=.7\textwidth,alt={Heat map of CLIFS and baseline method performance across datasets and text-length thresholds}]{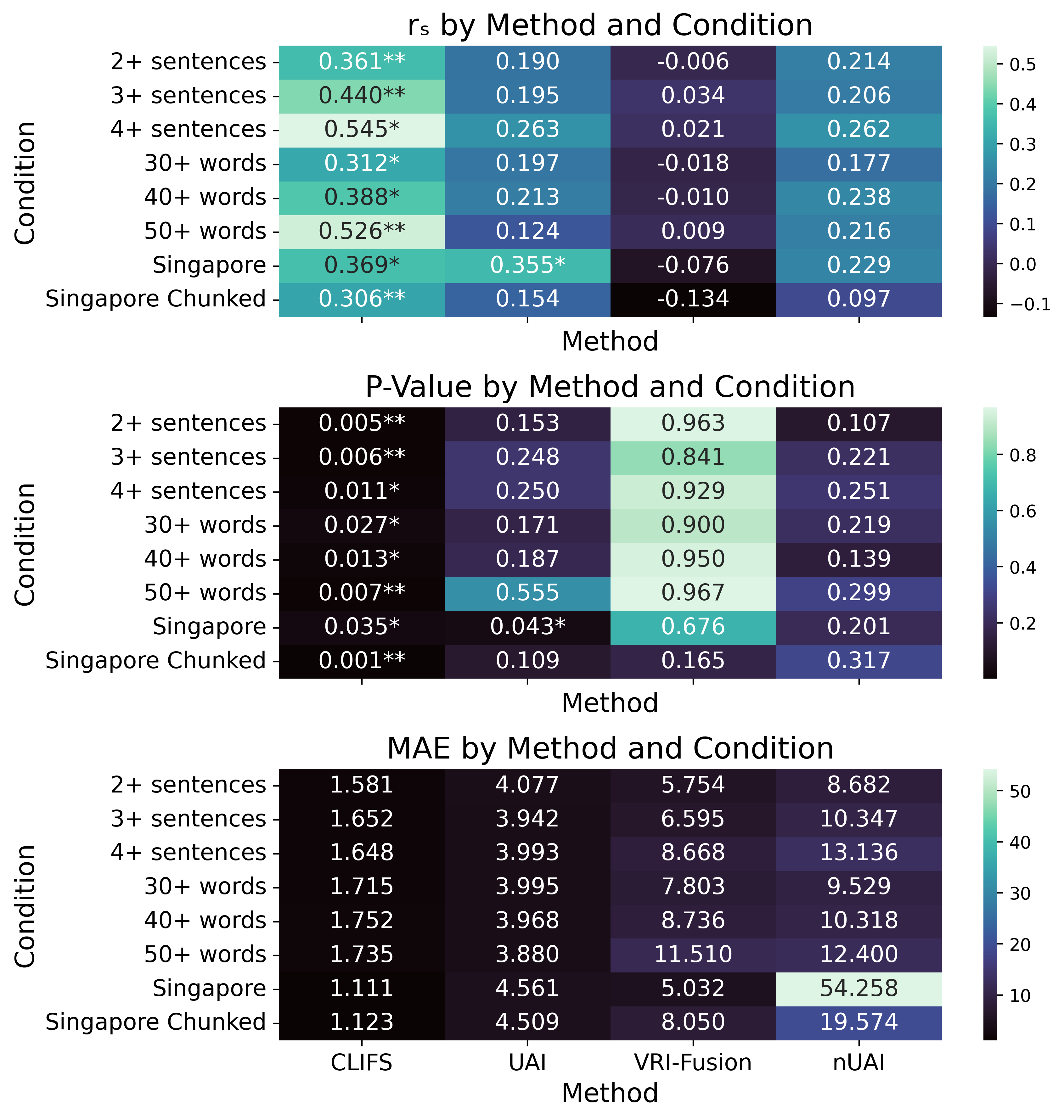} 

	\caption{\textbf{Performance of fusion-prediction methods across text-length and sampling conditions.}
        Heat maps show Spearman rank correlations ($r_s$; top), associated $p$ values (middle), and mean absolute error (MAE; bottom) between model scores and ground-truth identity fusion ratings for four methods (CLIFS, UAI, VRI-Fusion, and nUAI) across minimum-length and corpus conditions. Rows correspond to filtering thresholds based on the number of sentences or words per document in the MTurk corpus, as well as the Singapore corpus in original and chunked form. Higher $r_s$ and lower MAE indicate better agreement with true scores. Statistical significance is indicated by asterisks ($^*p<0.05$, $^{**}p<0.01$).}

	\label{fig:heatmap} 
\end{figure}

Figure~\ref{fig:heatmap} reports the performance of CLIFS and baseline methods (UAI, nUAI, and VRI-Fusion) across the Field and MTurk datasets, under multiple filtering conditions. Results are presented in terms of Spearman's rank correlation coefficient ($r_s$), statistical significance ($p$-values), and mean absolute error (MAE).  

Across all conditions, CLIFS achieved the highest correlations with ground-truth VIFS scores. In the MTurk dataset, CLIFS correlations ranged from $r_s = 0.31$--$0.55$ depending on filtering thresholds, all of which were statistically significant ($p < 0.05$; $0.005$--$0.027$). Correlation strength increased as higher minimum thresholds were enforced, indicating context improves performance. In contrast, none of the baseline methods produced significant correlations in any MTurk condition ($p = 0.107$--$0.967$). Similar results were observed in the Singapore field data: CLIFS again outperformed all baselines, yielding significant correlations both when using full transcripts ($r_s = 0.37$, $p = 0.035$) and when applying the chunked analysis ($r_s = 0.31$, $p = 0.001$).  

In all conditions but one, UAI did not achieve significance ($p = 0.109$--$0.555$). A significant correlation emerged only in the full transcript Singapore condition ($r_s = 0.36$; $p = 0.043$); however, this effect did not generalize to chunked data. Consistent with prior findings, UAI's reliance on z-scores renders it unreliable in settings with different identity fusion distributions (e.g., a highly fused sample of extremist manifestos)~\cite{wright2025clifs}. By contrast, VRI-Fusion ($p = 0.165$--$0.967$) and nUAI ($p = 0.107$--$0.317$) failed to achieve significance in any condition on either dataset.

MAE results tell the same story: CLIFS consistently achieved the lowest error across all conditions, by a wide margin. MAE for CLIFS ranged from $1.111$--$1.752$, compared to $3.88$--$4.561$ for UAI, $5.032$--$11.510$ for VRI-Fusion, and $8.682$--$54.258$ for nUAI.  

\subsubsection*{Divergent Pathways to Violence}

\begin{figure} 
	\centering
	\includegraphics[width=1\textwidth,alt={Histogram, KDE, and ECDF comparison of CLIFS scores for victim vs ideologue texts}]{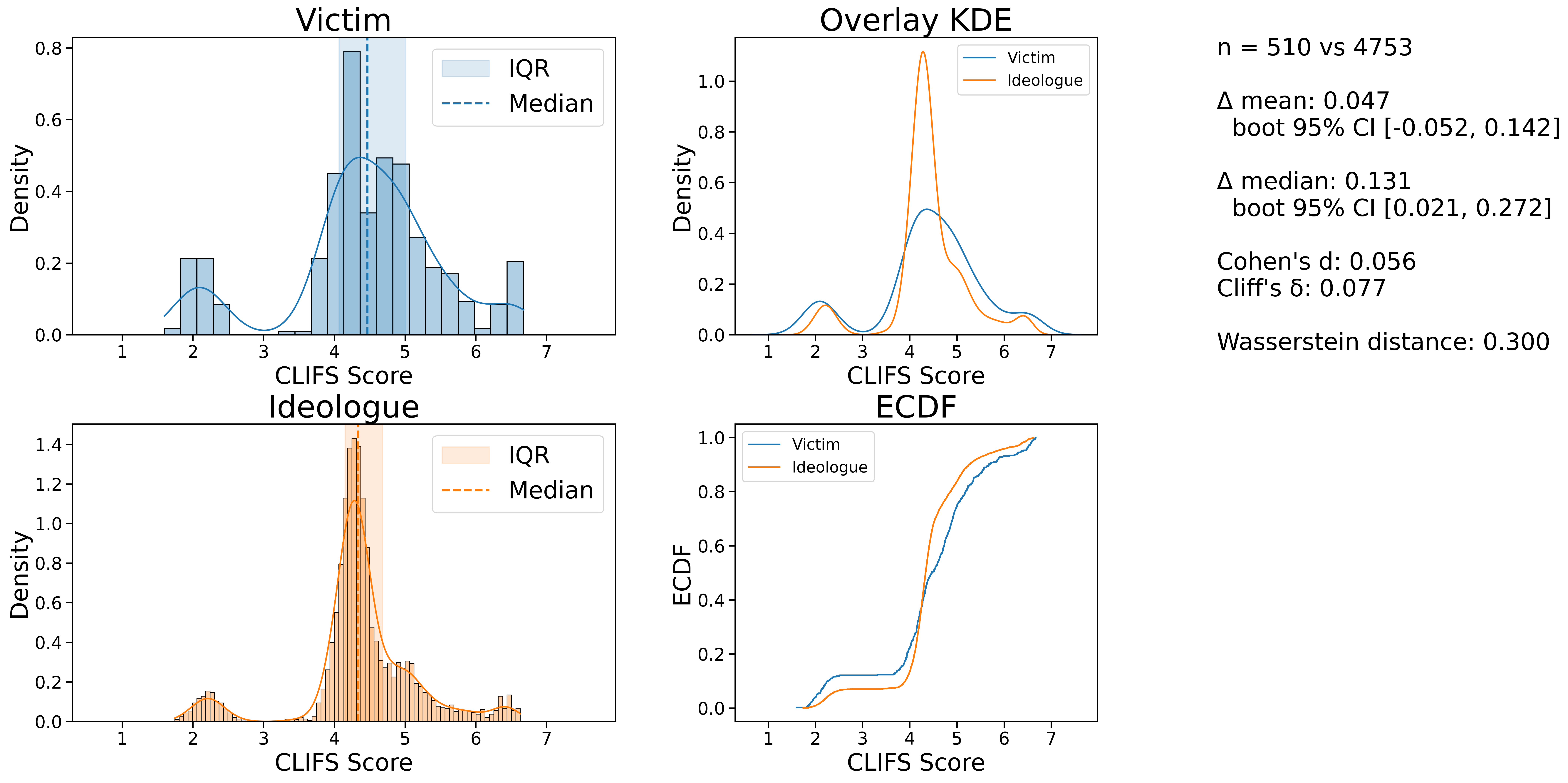} 
    \caption{
    \textbf{Distributional comparison of CLIFS scores by motivational framing.}
        CLIFS score distributions for victim- and ideologue-motivated texts are shown using density-normalized histograms and kernel density estimates (KDE), with interquartile ranges (IQR; shaded) and medians (dashed). Overlayed KDEs (top right) and empirical cumulative distribution functions (ECDF; bottom right) enable direct shape and distribution-wide comparison. Differences in central tendency and effect sizes (Victim $-$ Ideologue) are small, and the Wasserstein distance indicates only a modest global shift between distributions.}

	\label{fig:all_hist_kde} 
\end{figure}

\begin{figure}
	\centering
	\includegraphics[width=1\textwidth,alt={Four-panel histograms and KDEs for CLIFS component scores by motivational framing}]{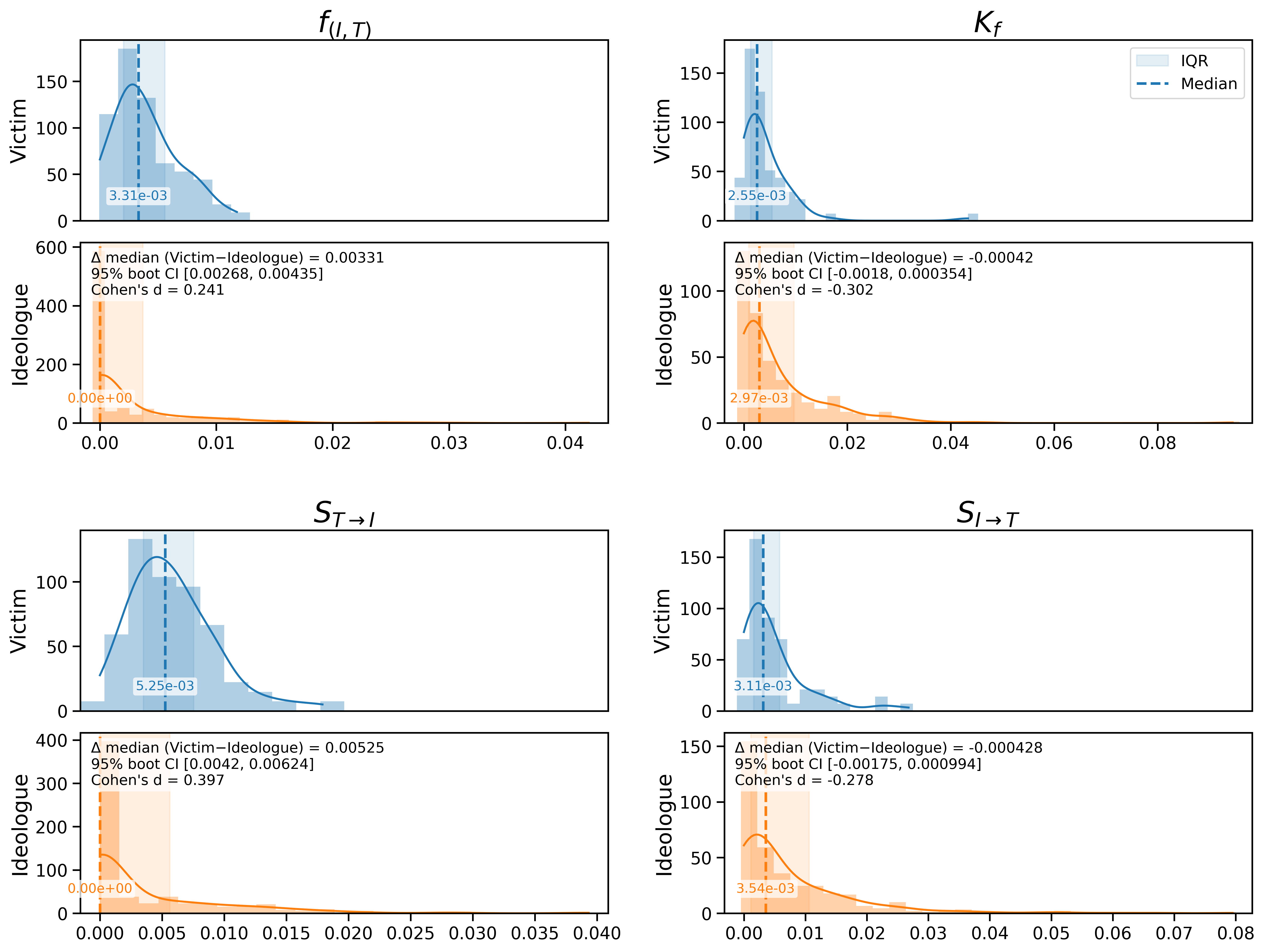} 

	\caption{\textbf{Distributions of CLIFS masked-LM implicit-metaphor scores for victim- and ideologue-motivated texts.}
        Histograms and kernel density estimates show the distributions of the four CLIFS components—fusion proximity $f_{(I,T)}$, fictive kinship $K_f$, fusion target-to-identity $S_{T \rightarrow I}$, and identity-to-target $S_{I \rightarrow T}$ directional scores—for text chunks labeled High Fusion by the coarse-grained CLIFS classifier. Texts classified as \emph{Victim} are shown in blue (top panels) and \emph{Ideologue} in orange (bottom panels). Dashed vertical lines indicate group medians, and shaded regions indicate interquartile ranges (IQRs). Numeric annotations report median values. Insets summarize median differences (Victim $-$ Ideologue) with 95\% bootstrap confidence intervals and Cohen's $d$.}

	\label{fig:2x2_hist_kde}
\end{figure}

Figure~\ref{fig:all_hist_kde} illustrates the CLIFS distributions across groups fall within a similar range, exhibit comparable central tendencies, and show only small differences ($\Delta$ mean: 0.047, 95\% CI: $[-0.052,0.142]$; $\Delta$ median: $0.131$, 95\% CI: $[0.021,0.272]$; Cohen's $d$: $0.056$; Cliff's $\delta$: $0.077$; Wasserstein distance: $0.3$), indicating broadly similar levels of identity fusion across pathways to violence---though marginally higher for the Victim class. To further examine differences within the fusion conceptual space, we restrict analysis to text segments labeled \emph{High Fusion} by the coarse-grained CLIFS classifier and visualize the implicit metaphor component features (Figure~\ref{fig:2x2_hist_kde}). Distinct patterns emerge. Victimhood-driven texts exhibit higher $f_{(I,T)}$ (Victim, median: $3.31 \times 10^{-3}$; Ideologue, median: $0$; 95\% CI: $[0.003,0.004]$; Cohen's $d$: $0.241$) and greater $S_{T \rightarrow I}$ (Victim, median: $5.25 \times 10^{-3}$; Ideologue, median: $0$; 95\% CI: $[0.004,0.006]$; Cohen's $d$: $0.397$), whereas $K_f$ (Victim, median: $2.55 \times 10^{-3}$; Ideologue, median: $2.97 \times 10^{-3}$; 95\% CI: $[-0.0018,0.0004]$; Cohen's $d$: $-0.302$) and $S_{I \rightarrow T}$ (Victim, median: $3.11 \times 10^{-3}$; Ideologue, median: $3.54 \times 10^{-3}$; 95\% CI: $[-0.002,0.001]$; Cohen's $d$: $-0.278$) show more similar distributions across groups, with comparatively less pronounced increases for ideologically motivated texts.

\subsection*{Discussion}

\subsubsection*{Theoretical and Practical Advances}

Cross-cultural evaluations show that CLIFS captures identity fusion across linguistic, cultural, and communicative contexts, generalizing to both out-of-domain datasets. This strengthens the case for CLIFS as a theoretically grounded and practically useful tool, not only for violence-risk prediction but also for advancing the broader study of identity fusion in naturalistic contexts, and at a much larger scale than was previously possible.

By contrast, baseline models failed to achieve significance in all conditions (except one for UAI), revealing the limitations of measures dependent on distributional assumptions or string-matching heuristics. Even UAI's limited success in the Singapore data required complete transcripts and a full dataset (i.e., UAI cannot operate on a single sample), as well as data properties aligned with its assumptions, illustrating the risk of metrics that cannot adapt to heterogeneous textual environments---a major limitation when studying extremist or online communication.

Moreover, contrary to our hypothesis that grievance-driven individuals would exhibit weaker identity fusion, both groups show comparable fusion patterns, with victimhood-driven texts exhibiting slightly higher overall fusion. These similarities are nonetheless expressed in distinct ways. Ideological fusion appears more strongly associated with fictive-kinship, conceptualizing the group in familial terms and aligning identity with shared collective semantics. Victimhood-driven individuals project self-views outward, construing others as similar to themselves (potentially without engaging divergent minds through metacognitive practice or by fusing to an abstraction other than a social group), alongside attenuated fictive-kinship dynamics.

This distinction refines understanding of variation in identity fusion expression, fusion between self and target can be potentially asymmetric and/or non-kin-based, producing distinct fusion-motivated pathways to violence. Our findings align with recent updates to Comprehensive Identity Fusion Theory (CIFT), which expand fusion beyond social groups to include values, ideologies, and abstract targets~\cite{SwannKleinGomez2024,GomezChinchillaVazquezEtAl2020}. CLIFS appears sensitive to variable constellations of fusion, reinforcing recent theoretical shifts toward broader models of fusion.

An important contribution of this work is clarifying the limitations of prior NLP approaches to fusion, particularly the VRI method---considered state-of-the-art in identity-fusion-based linguistic violence risk assessment~\cite{EbnerKavanaghWhitehouse2024,EbnerKavanaghWhitehouse2024b,EbnerKavanaghWhitehouse2022,EbnerKavanaghWhitehouse2022c,Oxford2023_MRSPresidentsMedal,Oxford2023_InfluentialResearch}. Both the original CLIFS paper~\cite{wright2025clifs} and the present study converge on the same conclusion: \textit{VRI does not provide a valid or generalizable measure of fusion or violence risk}.

Their method employs string-matching combined with manual review, justified by highlighting AI-based NLP error rates~\cite{EbnerKavanaghWhitehouse2024b}. Yet this hybrid design eliminates scalability and reproducibility, forcing analysts to hand-filter thousands of texts---an impossible standard for modern NLP, where corpora are often millions of samples. The authors themselves acknowledge ``a very high rate of false positives'' but fail to report them~\cite{EbnerKavanaghWhitehouse2024}, and instead apply unverified manual corrections that obscure the method's reliability and invalidates their results both when testing on in-domain and out-of-domain data~\cite{EbnerKavanaghWhitehouse2022c,EbnerKavanaghWhitehouse2024b,EbnerKavanaghWhitehouse2024}.

Crucially, the same manifestos used for feature derivation were reused for testing, introducing circularity and violating basic NLP standards requiring held-out validation~\cite{jm3}. Even under these favorable conditions, extensive manual filtering was needed to obtain reportable results~\cite{EbnerKavanaghWhitehouse2024,EbnerKavanaghWhitehouse2024b}. Additionally, prior work demonstrates that without artificial filtering, \textit{the VRI performs no better than chance on its own development data}~\cite{wright2025clifs}.

To their credit, the authors propose integrating machine learning through CASM Technology's Method52 \cite{EbnerKavanaghWhitehouse2024b}, yet this suggestion reflects a misunderstanding of both the tool and the problem space. Method52 is not a novel algorithm but a modular platform enabling \textit{non-technical researchers and analysts to assemble pipelines} and train task-specific classifiers \cite{SmithBartlettBuckHoneyman2017,settles2011closing,BCS_2017}; it cannot resolve underlying issues of construct or method validity. Although VRI is framed as a violence-risk method, it is fundamentally an identity-fusion-based metric \cite{EbnerKavanaghWhitehouse2024}. Without first establishing a valid measure of fusion, any derived risk framework compounds methodological uncertainty rather than reducing it. The original CLIFS paper showed that replacing the VRI-Fusion module with CLIFS metrics improved violence-risk prediction by over 240\% \cite{wright2025clifs}. This illustrates that establishing a valid measure of identity fusion is a necessary first step before attempting to build a violence-risk framework upon it. Without this foundation, plug-and-play architectures risk producing results that appear rigorous but lack connection to the psychological and social processes they aim to capture.

Furthermore, across two additional out-of-domain corpora, \textit{we again find no significant correlations between the VRI-Fusion method and ground-truth fusion scores}. Because VRI-Fusion requires extensive manual filtering of false positives prior to reporting results~\cite{EbnerKavanaghWhitehouse2024,EbnerKavanaghWhitehouse2024b,EbnerKavanaghWhitehouse2022,EbnerKavanaghWhitehouse2022c}, VRI success appears to result from post-hoc curation, rather than from scalable or generalizable measures of fusion or violence risk.

By contrast, CLIFS requires no manual filtering and achieves reliable, out-of-domain performance across multiple datasets. This demonstrates that CLIFS captures underlying cognitive-linguistic markers of fusion rather than dataset-specific artifacts. The contrast highlights a broader methodological point: scalable, reproducible measurement of fusion requires moving beyond lexicon-based heuristics toward theoretically grounded, computational methods.

Our findings also raise questions about the nature of threat in motivating violence among fused individuals. Existing models, such as the fusion-plus-threat model and the fusion-secure base hypothesis, emphasize that fusion predicts extreme pro-group behavior primarily under conditions of perceived threat~\cite{KleinBastian2023,SwannKleinGomez2024,EbnerKavanaghWhitehouse2022,Whitehouse2018}. Prior work stressed threats as key triggers of self-sacrifice~\cite{Whitehouse2018,EbnerKavanaghWhitehouse2024b}. However, prior work shows that even non-lethal and symbolic outgroup threats (e.g., rival football team fans) can activate violent readiness among highly fused individuals~\cite{Whitehouse2018}. In light of our results, it may also be social (e.g., ostracism, exile) or conceptual (e.g., the symbolic ``death'' of an idealized self). Our results likewise suggest that such non-existential threats can suffice to catalyze violence, particularly when fusion takes asymmetric or non-kin-based forms. Furthermore, self-verification (others seeing us as we see ourselves) has been shown to foster fusion; perhaps the rejection of self (by others) triggers violence in already fused individuals~\cite{gomez2024feeling}.

CLIFS thus provides a novel means to quantify, differentiate, and observe qualitative differences across expressions of fusion, opening the door to systematic investigation of the complex dynamics and feedback loops that shape how different types of threat interact with different forms of fusion to motivate violence.

\subsection*{Conclusion}

Growing concerns about online radicalization and real-world political violence present stakeholders such as security and defense institutions and social media companies with a challenge: how to determine whether online comments by individuals anywhere on the ideological spectrum suggest they are approaching a radicalization threshold that is likely to lead them to engage in extreme actions that will harm others. Models that rely on the analysis of topics are insufficient because simply discussing a topic is not a sign of radicalization. Previous research has shown that a high level of identity fusion appears to be a necessary psychological predictor of a willingness to endorse or engage in extreme acts of violence against ideological others~\cite{SwannKleinGomez2024,wright2025clifs,AshokkumarSwann2023,BestaGomezVazquez2014,WhitehouseMcQuinnBuhrmesterSwann2014,SwannGomezDovidioHartJetten2010}. Targeting identity fusion is therefore a valid and reasonable way to reduce the search space for the potentially violent ``needle'' in the \textit{massive} ``haystack'' of online media data.

The CLIFS system presented here demonstrates far greater accuracy in predicting identity fusion (and violence~\cite{wright2025clifs}) than previous systems as well as the technical ability to scale. This opens up new opportunities for social media organizations to create interventions that address the very real problem of online radicalization without relying on problematic content moderation or censorship policies that all too often have led to greater polarization and political radicalization~\cite{laneetal2021}. Instead of analyzing superficial factors such as political preference or word frequency, CLIFS provides an effective way to detect the actual motivational factors driving extremism at the underlying psychological level.

\begin{table}[htbp]
\centering
\caption{\textbf{Violent Manifestos Dataset.}
Catalog of extremist and mass-violence texts analyzed in this study, listed by author and associated incident or work. For each document, the file format and a high-level classification of motivational framing (\emph{Ideologue} or \emph{Victim}) are reported. These identifiers enable independent researchers to locate the same source materials used in the analyses without redistributing harmful content.}

\small
\setlength{\tabcolsep}{4pt}
\renewcommand{\arraystretch}{1.05}
\label{tab:manifestos}
\begin{tabularx}{\linewidth}{|p{4.2cm}|X|c|c|}
\hline
\textbf{Author} & \textbf{Incident / Work} & \textbf{File Type} & \textbf{Motivation} \\
\hline
Payton S. Gendron & Buffalo Shooter & PDF & Ideologue \\
\hline
Theodore Kaczynski & Unabomber & PDF & Ideologue \\
\hline
Kyle Odom & Idaho Pastor Shooting & PDF & Ideologue \\
\hline
Seung-hui Cho & Virginia Tech Shooting & PDF & Victim \\
\hline
Christopher Dorner & Christopher Dorner shootings & DOCX & Victim \\
\hline
Pekka-Eric Auvinen & Jokela High School Shooting & TXT & Ideologue \\
\hline
Jerad and Amanda Miller & Las Vegas Shooting & MP4 & Ideologue \\
\hline
Eric Harris and Dylan Klebold & Columbine High School Massacre & DOCX & Victim \\
\hline
Matthew Livelsberger & Cybertruck Bombing & DOCX & Ideologue \\
\hline
Anders Behring Breivik & Norway Attacks & DJVU TXT & Ideologue \\
\hline
Elliot Rodger & Isla Vista Killing & DJVU TXT & Victim \\
\hline
Dylann Roof & Charleston Church Shooting & DJVU TXT & Ideologue \\
\hline
Brenton Tarrant & Christchurch Mosque Shooting & DJVU TXT & Ideologue \\
\hline
Stephan Balliet & Halle Synagogue Shooting & DJVU TXT & Ideologue \\
\hline
John Earnest & Poway Synagogue Shooting & DJVU TXT & Ideologue \\
\hline
Patrick Crusius & El Paso Walmart Shooting & DJVU TXT & Ideologue \\
\hline
Adolf Hitler & \emph{Mein Kampf} & DJVU TXT & Ideologue \\
\hline
Sayyid Qutb & \emph{Milestones} & DJVU TXT & Ideologue \\
\hline
\end{tabularx}
\end{table}

\clearpage
\bibliographystyle{unsrtnat}
\bibliography{references}

@inproceedings{wright2025clifs,
    title = "Cognitive Linguistic Identity Fusion Score ({CLIFS}): A Scalable {C}ognition{-}{I}nformed Approach to Quantifying Identity Fusion from Text",
    author = "Wright, Devin R.  and
      An, Jisun  and
      Ahn, Yong-Yeol",
    editor = "Christodoulopoulos, Christos  and
      Chakraborty, Tanmoy  and
      Rose, Carolyn  and
      Peng, Violet",
    booktitle = "Proceedings of the 2025 Conference on Empirical Methods in Natural Language Processing",
    month = nov,
    year = "2025",
    address = "Suzhou, China",
    publisher = "Association for Computational Linguistics",
    url = "https://aclanthology.org/2025.emnlp-main.588/",
    doi = "10.18653/v1/2025.emnlp-main.588",
    pages = "11643--11673",
    ISBN = "979-8-89176-332-6"
}

@misc{CLIFS_github,
  author       = {Devin R. Wright},
  title        = {CLIFS: Cognitive Linguistic Identity Fusion Score, GitHub Repository},
  year         = {2025},
  howpublished = {\url{https://github.com/DevinW-sudo/CLIFS}}
}

@article{CardChangEtAl2022,
  author    = {Dallas Card and Serina Chang and Chris Becker and Julia Mendelsohn and Rob Voigt and Leah Boustan and Ran Abramitzky and Dan Jurafsky},
  title     = {Computational analysis of 140 years of US political speeches reveals more positive but increasingly polarized framing of immigration},
  journal   = {Proceedings of the National Academy of Sciences},
  year      = {2022},
  volume    = {119},
  number    = {31},
  pages     = {e2120510119},
  doi       = {10.1073/pnas.2120510119},
  url       = {https://www.pnas.org/doi/abs/10.1073/pnas.2120510119}
}

@misc{modernbert,
      title={Smarter, Better, Faster, Longer: A Modern Bidirectional Encoder for Fast, Memory Efficient, and Long Context Finetuning and Inference}, 
      author={Benjamin Warner and Antoine Chaffin and Benjamin Clavié and Orion Weller and Oskar Hallström and Said Taghadouini and Alexis Gallagher and Raja Biswas and Faisal Ladhak and Tom Aarsen and Nathan Cooper and Griffin Adams and Jeremy Howard and Iacopo Poli},
      year={2024},
      eprint={2412.13663},
      archivePrefix={arXiv},
      primaryClass={cs.CL},
      url={https://arxiv.org/abs/2412.13663}, 
}

@article{AshokkumarPennebaker2022,
  author  = {Ashwini Ashokkumar and James W. Pennebaker},
  title   = {Tracking group identity through natural language within groups},
  journal = {PNAS Nexus},
  year    = {2022},
  volume  = {1},
  number  = {2},
  pages   = {pgac022},
  doi     = {10.1093/pnasnexus/pgac022},
  url     = {https://doi.org/10.1093/pnasnexus/pgac022}
}

@techreport{pennebaker2015development,
  author       = {Pennebaker, James W. and Boyd, Ryan and Jordan, Kayla and Blackburn, Kate},
  title        = {The development and psychometric properties of LIWC2015},
  institution  = {University of Texas at Austin},
  address      = {Austin, TX},
  year         = {2015},
  type         = {Commissioned report},
  pages        = {27},
  doi          = {10.15781/T29G6Z},
  url          = {https://doi.org/10.15781/T29G6Z}
}

@article{GomezBrooksEtAl2011,
  author  = {Angel Gómez and Matthew Brooks and Michael Buhrmester and Alexandra Vázquez and Jolanda Jetten and William Swann},
  title   = {On the Nature of Identity Fusion: Insights Into the Construct and a New Measure},
  journal = {Journal of Personality and Social Psychology},
  year    = {2011},
  volume  = {100},
  pages   = {918--933},
  doi     = {10.1037/a0022642}
}

@article{EbnerKavanaghWhitehouse2024,
  author    = {Julia Ebner and Christopher Kavanagh and Harvey Whitehouse},
  title     = {Assessing Violence Risk among Far-Right Extremists: A New Role for Natural Language Processing},
  journal   = {Terrorism and Political Violence},
  year      = {2024},
  volume    = {36},
  number    = {7},
  pages     = {944--961},
  doi       = {10.1080/09546553.2023.2236222},
  url       = {https://www.tandfonline.com/doi/full/10.1080/09546553.2023.2236222}
}

@article{EbnerKavanaghWhitehouse2024b,
  author    = {Julia Ebner and Christopher Kavanagh and Harvey Whitehouse},
  title     = {Measuring socio-psychological drivers of extreme violence in online terrorist manifestos: an alternative linguistic risk assessment model},
  journal   = {Journal of Policing, Intelligence and Counter Terrorism},
  year      = {2024},
  volume    = {19},
  number    = {2},
  pages     = {125--143},
  doi       = {10.1080/18335330.2023.2246982},
  url       = {https://www.tandfonline.com/doi/full/10.1080/18335330.2023.2246982}
}

@article{EbnerKavanaghWhitehouse2022c,
  author    = {Julia Ebner and Christopher Kavanagh and Harvey Whitehouse},
  title     = {The QAnon Security Threat: A Linguistic Fusion-Based Violence Risk Assessment},
  journal   = {Perspectives on Terrorism},
  year      = {2022},
  volume    = {16},
  number    = {6},
  pages     = {62--86},
  url       = {https://www.jstor.org/stable/27185092},
  publisher = {Terrorism Research Initiative}
}

@article{EbnerKavanaghWhitehouse2022,
  author    = {Julia Ebner and Chris Kavanagh and Harvey Whitehouse},
  title     = {Is There a Language of Terrorists? A Comparative Manifesto Analysis},
  journal   = {Studies in Conflict \& Terrorism},
  year      = {2022},
  volume    = {0},
  number    = {0},
  pages     = {1--27},
  doi       = {10.1080/1057610X.2022.2109244},
  url       = {https://doi.org/10.1080/1057610X.2022.2109244},
  publisher = {Routledge}
}

@incollection{SwannKleinGomez2024,
  author    = {William B. Swann and Jack W. Klein and {Ángel Gómez}},
  title     = {Comprehensive identity fusion theory (CIFT): New insights and a revised theory},
  booktitle = {Advances in Experimental Social Psychology},
  year      = {2024},
  volume    = {70},
  pages     = {275--332},
  publisher = {Elsevier},
  doi       = {10.1016/bs.aesp.2024.03.003},
  url       = {https://linkinghub.elsevier.com/retrieve/pii/S0065260124000157},
  isbn      = {978-0-443-29428-0}
}

@article{FangXie2022,
  author  = {Hongchao Fang and Pengtao Xie},
  title   = {An End-to-End Contrastive Self-Supervised Learning Framework for Language Understanding},
  journal = {Transactions of the Association for Computational Linguistics},
  year    = {2022},
  volume  = {10},
  pages   = {1324--1340},
  doi     = {10.1162/tacl\_a\_00521},
  url     = {https://doi.org/10.1162/tacl\_a\_00521}
}

@inproceedings{ZhangMiZhouEtAl2024,
  author    = {Dawei Zhang and Rongxin Mi and Peiyao Zhou and Dawei Jin and Manman Zhang and Tianhang Song},
  title     = {Data Augmentation for Imbalanced Text Classification Using Large Language Models},
  booktitle = {Proceedings of the 2024 5th International Seminar on Artificial Intelligence, Networking and Information Technology (AINIT)},
  year      = {2024},
  month     = {March},
  pages     = {1006--1010},
  doi       = {10.1109/AINIT61980.2024.10581735},
  url       = {https://ieeexplore.ieee.org/document/10581735/?arnumber=10581735},
  publisher = {IEEE}
}

@book{Bird_Natural_Language_Processing_2009,
author = {Bird, Steven and Klein, Ewan and Loper, Edward},
isbn = {9780596516499},
month = jun,
publisher = {O'Reilly Media, Inc.},
title = {{Natural Language Processing with Python: Analyzing Text with the Natural Language Toolkit}},
url = {https://www.nltk.org/book/},
year = {2009}
}

@inproceedings{reimers-2019-sentence-bert,
  title = "Sentence-BERT: Sentence Embeddings using Siamese BERT-Networks",
  author = "Reimers, Nils and Gurevych, Iryna",
  booktitle = "Proceedings of the 2019 Conference on Empirical Methods in Natural Language Processing",
  month = "11",
  year = "2019",
  publisher = "Association for Computational Linguistics",
  url = "https://arxiv.org/abs/1908.10084",
}

@article{song2020mpnet,
  title={Mpnet: Masked and permuted pre-training for language understanding},
  author={Song, Kaitao and Tan, Xu and Qin, Tao and Lu, Jianfeng and Liu, Tie-Yan},
  journal={Advances in neural information processing systems},
  volume={33},
  pages={16857--16867},
  year={2020}
}

@article{GomezChinchillaVazquezEtAl2020,
  author    = {Ángel Gómez and Juana Chinchilla and Alexandra Vázquez and Lucía López‐Rodríguez and Borja Paredes and Mercedes Martínez},
  title     = {Recent advances, misconceptions, untested assumptions, and future research agenda for identity fusion theory},
  journal   = {Social \& Personality Psychology Compass},
  year      = {2020},
  volume    = {14},
  number    = {6},
  pages     = {1--15},
  doi       = {10.1111/spc3.12531},
  url       = {https://doi.org/10.1111/spc3.12531}
}

@article{KleinBastian2023,
  author       = {Klein, Jack W. and Bastian, Brock},
  title        = {The Fusion-Secure Base Hypothesis},
  journal      = {Personality and Social Psychology Review},
  volume       = {27},
  number       = {2},
  pages        = {107--127},
  year         = {2023},
  month        = may,
  publisher    = {SAGE Publications Inc},
  doi          = {10.1177/10888683221100883},
  url          = {https://doi.org/10.1177/10888683221100883},
  issn         = {1088-8683},
}

@article{Whitehouse2018,
  author       = {Whitehouse, Harvey},
  title        = {Dying for the Group: Towards a General Theory of Extreme Self-Sacrifice},
  journal      = {Behavioral and Brain Sciences},
  volume       = {41},
  pages        = {e192},
  year         = {2018},
  month        = jan,
  publisher    = {Cambridge University Press},
  doi          = {10.1017/S0140525X18000249},
  url          = {https://doi.org/10.1017/S0140525X18000249},
  issn         = {0140-525X, 1469-1825},
}

@article{gomez2024feeling,
  title={Feeling understood fosters identity fusion.},
  author={G{\'o}mez, Angel and V{\'a}zquez, Alexandra and Alba, Beatriz and Blanco, Laura and Chinchilla, Juana and Chiclana, Sandra and Swann Jr, William B},
  journal={Journal of Personality and Social Psychology},
  year={2024},
  publisher={American Psychological Association}
}

@misc{BCS_2017,
  author       = {{BCS, The Chartered Institute for IT}},
  title        = {How politicians are learning from social media},
  year         = {2017},
  howpublished = {\url{https://www.bcs.org/articles-opinion-and-research/how-politicians-are-learning-from-social-media}},
  note         = {Accessed Aug 2025}
}

@inproceedings{settles2011closing,
  title={Closing the loop: Fast, interactive semi-supervised annotation with queries on features and instances},
  author={Settles, Burr},
  booktitle={Proceedings of the 2011 conference on empirical methods in natural language processing},
  pages={1467--1478},
  year={2011}
}

@techreport{SmithBartlettBuckHoneyman2017,
  author = {Smith, Josh and Bartlett, Jamie and Buck, David and Honeyman, Matthew},
  title = {Online Support: Investigating the role of public online forums in mental health},
  institution = {Centre for the Analysis of Social Media (CASM) at Demos, in partnership with The King’s Fund},
  year = {2017},
  month = apr,
  publisher = {Demos},
  note = {Funded by the Wellcome Trust. Open access.},
  url = {https://www.demos.co.uk/wp-content/uploads/2017/04/Online-Support-Demos-report.pdf}
}

@Book{jm3,
  author =       "Daniel Jurafsky and James H. Martin",
  title =        "Speech and Language Processing: An Introduction to Natural Language Processing, 
  		  Computational Linguistics, and Speech Recognition,
		   with Language Models",
  year =         "2025",
  url = {https://web.stanford.edu/~jurafsky/slp3/},
  note = "Online manuscript released August 24, 2025",
  edition =         "3rd",
  }

@misc{Oxford2023_MRSPresidentsMedal,
  author       = {{University of Oxford, School of Anthropology \& Museum Ethnography}},
  title        = {Work to predict extreme violence amongst online users wins MRS President’s Medal},
  year         = {2023},
  url = {https://www.anthro.ox.ac.uk/article/work-predict-extreme-violence-amongst-online-users-wins-mrs-presidents-medal},
  language     = {English}
}

@misc{Oxford2023_InfluentialResearch,
  author       = {{University of Oxford, School of Anthropology \& Museum Ethnography}},
  title        = {Influential and life-changing research recognised},
  year         = {2023},
  note         = {Accessed Aug 2025},
  url = {https://www.anthro.ox.ac.uk/article/influential-and-life-changing-research-recognised},
  language     = {English}
}

@article{SwannJettenGomezWhitehouseBastian2012,
  author  = {William B. Swann and Jolanda Jetten and Ángel Gómez and Harvey Whitehouse and Brock Bastian},
  title   = {When Group Membership Gets Personal: A Theory of Identity Fusion},
  journal = {Psychological Review},
  year    = {2012},
  volume  = {119},
  number  = {3},
  pages   = {441--456},
  doi     = {10.1037/a0028589},
  url     = {https://doi.org/10.1037/a0028589},
  note    = {Epub 2012 May 28}
}

@article{SwannGomezEtAl2009,
  author    = {William B. Swann and {Ángel Gómez} and D. Conor Seyle and J. Francisco Morales and Carmen Huici},
  title     = {Identity fusion: The interplay of personal and social identities in extreme group behavior},
  journal   = {Journal of Personality and Social Psychology},
  year      = {2009},
  volume    = {96},
  number    = {5},
  pages     = {995--1011},
  doi       = {10.1037/a0013668},
  url       = {https://doi.apa.org/doi/10.1037/a0013668}
}

@article{Kristinsdottir_Ebner_Whitehouse_2025, title={Extreme overvalued beliefs and identities: revisiting the drivers of violent extremism}, volume={16}, url={http://dx.doi.org/10.3389/fpsyg.2025.1556919}, DOI={10.3389/fpsyg.2025.1556919}, journal={Frontiers in Psychology}, publisher={Frontiers Media SA}, author={Kristinsdottir, Kolbrun and Ebner, Julia and Whitehouse, Harvey}, year={2025}, month=mar }

@misc{Hooda_2025,
author = {Khyati Hoota},
title = {55 Facebook Stats for your Social Media Strategy [2025]},
url={https://keywordseverywhere.com/blog/facebook-stats/}, year={2025}, month=sep, language={en-US} }

@article{laneetal2021,
  author       = {Justin E. Lane and
                  Kevin McCaffree and
                  F. LeRon Shults},
  title        = {Is radicalization reinforced by social media censorship?},
  journal      = {CoRR},
  volume       = {abs/2103.12842},
  year         = {2021},
  url          = {https://arxiv.org/abs/2103.12842},
  eprinttype    = {arXiv},
  eprint       = {2103.12842},
  bibsource    = {dblp computer science bibliography, https://dblp.org}
}

@article{Goldstein2023UnderstandingT,
  title={Understanding the (In)Effectiveness of Content Moderation: A Case Study of Facebook in the Context of the U.S. Capitol Riot},
  author={Ian Goldstein and Laura Edelson and Damon McCoy and Tobias Lauinger},
  journal={ArXiv},
  year={2023},
  volume={abs/2301.02737},
  url={https://api.semanticscholar.org/CorpusID:255546156}
}

@article{doosje1995perceived,
  title={Perceived intragroup variability as a function of group status and identification},
  author={Doosje, Bertjan and Ellemers, Naomi and Spears, Russell},
  journal={Journal of experimental social psychology},
  volume={31},
  number={5},
  pages={410--436},
  year={1995},
  publisher={Elsevier}
}

@misc{postmes2013additional,
  title={Additional recommendations for measuring social identification},
  author={Postmes, Tom and Haslam, S Alex and Jans, Lise},
  year={2013}
}

@article{AshokkumarSwann2023,
  author    = {Ashwini Ashokkumar and William B. Swann},
  title     = {Restoring Honor by Slapping or Disowning the Daughter},
  journal   = {Personality and Social Psychology Bulletin},
  year      = {2023},
  volume    = {49},
  number    = {6},
  pages     = {823--836},
  doi       = {10.1177/01461672221079106},
  url       = {https://doi.org/10.1177/01461672221079106}
}

@article{BestaGomezVazquez2014,
  author    = {Tomasz Besta and {\'A}ngel G{\'o}mez and Alexandra V{\'a}zquez},
  title     = {Readiness to Deny Group's Wrongdoing and Willingness to Fight for Its Members: The Role of Poles' Identity Fusion with the Country and Religious Group},
  journal   = {Current Issues in Personality Psychology},
  year      = {2014},
  volume    = {2},
  number    = {1},
  pages     = {49--55},
  doi       = {10.5114/cipp.2014.43101},
  url       = {https://doi.org/10.5114/cipp.2014.43101}
}

@article{WhitehouseMcQuinnBuhrmesterSwann2014,
  author    = {Harvey Whitehouse and Brian McQuinn and Michael D. Buhrmester and William B. Swann Jr.},
  title     = {Brothers in arms: Libyan revolutionaries bond like family},
  journal   = {Proceedings of the National Academy of Sciences of the United States of America},
  year      = {2014},
  volume    = {111},
  number    = {50},
  pages     = {17783--17785},
  doi       = {10.1073/pnas.1416284111},
  url       = {https://www.pnas.org/doi/10.1073/pnas.1416284111},
  note      = {Epub 2014 Nov 10}
}

@article{SwannGomezDovidioHartJetten2010,
  author    = {William B. Swann and {\'A}ngel G{\'o}mez and John F. Dovidio and Sonia Hart and Jolanda Jetten},
  title     = {Dying and Killing for One's Group: Identity fusion moderates responses to intergroup versions of the trolley problem},
  journal   = {Psychological Science},
  year      = {2010},
  month     = {August},
  volume    = {21},
  number    = {8},
  pages     = {1176--1183},
  doi       = {10.1177/0956797610376656},
  url       = {https://www.jstor.org/stable/41062349},
  note      = {Epub 2010 Jul 9}
}

@misc{python_docx,
  title = {python-docx: a Python library for creating and updating Microsoft Word (.docx) files.},
  author = {Steve Canny and \textit{et al.}},
  url = {https://python-docx.readthedocs.io/}
}

@misc{pypdf,
 title         = {The {pypdf} library},
 author        = {Mathieu Fenniak and
                  Matthew Stamy and
                  pubpub-zz and
                  Martin Thoma and
                  Matthew Peveler and
                  exiledkingcc and {pypdf Contributors}},
 year          = {2024},
 url           = {https://pypi.org/project/pypdf/},
 note          = {See https://pypdf.readthedocs.io/en/latest/meta/CONTRIBUTORS.html for all contributors}
}

@inproceedings{smith2007overview,
  title={An overview of the Tesseract OCR engine},
  author={Smith, Ray},
  booktitle={Ninth international conference on document analysis and recognition (ICDAR 2007)},
  volume={2},
  pages={629--633},
  year={2007},
  organization={IEEE}
}

@misc{pytesseract,
  title        = {pytesseract: Python wrapper for Google Tesseract OCR},
  author       = {Samuel Hoffstaetter and \textit{et al.}},
  year         = {2025},
  url          = {https://github.com/madmaze/pytesseract},
  note         = {Version 0.3.13},
  license      = {Apache-2.0},
}

@misc{pymupdf,
  title={PyMuPDF},
  author={Artifex},
  url={https://pymupdf.readthedocs.io/}
}

@misc{vosk,
  title   = {Vosk Speech Recognition Toolkit: Offline speech recognition API for Android, iOS, Raspberry Pi and servers with Python, Java, C\# and Node},
  author  = {Shmyrev, Nickolay V. and \textit{et al.}},
  year    = {2020},
  url     = {https://github.com/alphacep/vosk-api}
}

@inproceedings{povey2011kaldi,
  title={The Kaldi speech recognition toolkit},
  author={Povey, Daniel and Ghoshal, Arnab and Boulianne, Gilles and Burget, Lukas and Glembek, Ondrej and Goel, Nagendra and Hannemann, Mirko and Motlicek, Petr and Qian, Yanmin and Schwarz, Petr and others},
  booktitle={IEEE 2011 workshop on automatic speech recognition and understanding},
  volume={1},
  pages={5--1},
  year={2011},
  organization={Hawaii}
}

@inproceedings{chen2021gigaspeech,
  title={GigaSpeech: An evolving, multi-domain ASR corpus with 10,000 hours of transcribed audio},
  author={Chen, Guoguo and Chai, Shuzhou and Wang, Guanbo and Du, Jiayu and Zhang, Wei Qiang and Weng, Chao and Su, Dan and Povey, Daniel and Trmal, Jan and Zhang, Junbo and others},
  booktitle={22nd Annual Conference of the International Speech Communication Association, INTERSPEECH 2021},
  pages={4376--4380},
  year={2021},
  organization={International Speech Communication Association}
}

@misc{methods,
  note = {Materials and methods are available as supplementary material},
}

@article{Waskom2021,
    doi = {10.21105/joss.03021},
    url = {https://doi.org/10.21105/joss.03021},
    year = {2021},
    publisher = {The Open Journal},
    volume = {6},
    number = {60},
    pages = {3021},
    author = {Michael L. Waskom},
    title = {seaborn: statistical data visualization},
    journal = {Journal of Open Source Software}
 }

@inproceedings{seabold2010statsmodels,
  title={statsmodels: Econometric and statistical modeling with python},
  author={Seabold, Skipper and Perktold, Josef},
  booktitle={9th Python in Science Conference},
  year={2010},
}

@article{2020SciPy-NMeth,
  author  = {Virtanen, Pauli and Gommers, Ralf and Oliphant, Travis E. and
            Haberland, Matt and Reddy, Tyler and Cournapeau, David and
            Burovski, Evgeni and Peterson, Pearu and Weckesser, Warren and
            Bright, Jonathan and {van der Walt}, St{\'e}fan J. and
            Brett, Matthew and Wilson, Joshua and Millman, K. Jarrod and
            Mayorov, Nikolay and Nelson, Andrew R. J. and Jones, Eric and
            Kern, Robert and Larson, Eric and Carey, C J and
            Polat, {\.I}lhan and Feng, Yu and Moore, Eric W. and
            {VanderPlas}, Jake and Laxalde, Denis and Perktold, Josef and
            Cimrman, Robert and Henriksen, Ian and Quintero, E. A. and
            Harris, Charles R. and Archibald, Anne M. and
            Ribeiro, Ant{\^o}nio H. and Pedregosa, Fabian and
            {van Mulbregt}, Paul and {SciPy 1.0 Contributors}},
  title   = {{{SciPy} 1.0: Fundamental Algorithms for Scientific
            Computing in Python}},
  journal = {Nature Methods},
  year    = {2020},
  volume  = {17},
  pages   = {261--272},
  adsurl  = {https://rdcu.be/b08Wh},
  doi     = {10.1038/s41592-019-0686-2},
}

\section*{Acknowledgments}
We would like to thank Jisun An for her feedback on early visualizations. We used generative AI tools during manuscript preparation and software development. For writing, we used AI across multiple sections to assist with language (e.g., improving clarity, conciseness, and phrasing). Another primary use was to reduce text to meet word limits; we reviewed and edited all AI-assisted output. For code, we used AI to generate low-novelty boilerplate and templates (e.g., pipeline scaffolding for data extraction from multiple document types). We take full responsibility for all content.

\paragraph*{Funding:}
This work was supported by CulturePulse, which provided financial support to all authors.

\paragraph*{Author contributions:}
D.R.W.: Conceptualization; Methodology; Investigation; Project administration; Data curation; Software; Formal analysis; Visualization; Validation; Writing -- original draft; Writing -- review \& editing.
J.E.L.: Conceptualization; Methodology; Investigation; Project administration; Resources, Data curation; Supervision; Validation; Writing -- original draft; Writing -- review \& editing.
F.L.S.: Supervision, Writing -- review \& editing.

\paragraph*{Competing interests:}
D.R.W. is a research scientist intern at CulturePulse. J.E.L. is the co-founder and Chief Executive Officer of CulturePulse. F.L.S. is the co-founder and Chief Research Officer of CulturePulse.

\paragraph*{Data and materials availability:}

The human-subjects data analyzed in this study were collected as part of approved research conducted under the University of Oxford School of Anthropology and Museum Ethnography ethics review process. Formal reference numbers were not issued for departmental anthropology ethics approvals at the time of data collection.

Due to ethical, legal, and safety considerations, the human-subjects datasets (field and MTurk data) cannot be made publicly available because they contain personally identifiable and sensitive information. De-identified data supporting the findings of this study are available to qualified researchers upon reasonable request, subject to approval and a data use agreement.

The corpus of extremist and mass-violence texts analyzed in this study is documented in Table \ref{tab:manifestos}, which provides identifying information for each document. These materials are publicly available from their original sources but are not redistributed here due to the harmful nature of the content. The identifiers in Table \ref{tab:manifestos} enable independent researchers to locate the exact documents used in this study.

The CLIFS scoring framework, trained models, and software used in this study (version 0.1.0) are publicly available at \url{https://github.com/DevinW-sudo/CLIFS}.

The analysis and visualization code used in this study is publicly available via the Open Science Framework (OSF, \url{https://osf.io/cas6d/overview}). CLIFS includes a full implementation of both nUAI and UAI, and a complete Python implementation of the VRI is also provided on OSF. However, several word and phrase lists required by the VRI contain harmful or potentially harmful content. To avoid distributing such material beyond what is strictly necessary, these lists have been left empty in the publicly shared implementation. The original lists can be accessed in the original R implementation of the VRI, where they are included as text within a Microsoft Word document provided in the supplementary materials of one of the VRI papers~\cite{EbnerKavanaghWhitehouse2024b}.

\subsection*{Supplementary materials}
Materials and Methods\\
References \textit{(40-51)}\\ 

\newpage

\renewcommand{\thefigure}{S\arabic{figure}}
\renewcommand{\thetable}{S\arabic{table}}
\renewcommand{\theequation}{S\arabic{equation}}
\renewcommand{\thepage}{S\arabic{page}}
\setcounter{figure}{0}
\setcounter{table}{0}
\setcounter{equation}{0}
\setcounter{page}{1}
\begin{center}
\section*{Supplementary Materials}

Devin~R.~Wright$^{\ast}$,
Justin~E.~Lane,
F.~LeRon~Shults\\
\small$^\ast$Corresponding author. Email: devrwrig@iu.edu\\
\end{center}

\subsubsection*{This PDF file includes:}
Materials and Methods\\
\newpage
\subsection*{Materials and Methods}

\subsubsection*{CLIFS Details}
\label{clifs_details}

Prior work demonstrated that masked-LM-based methods can effectively detect implicit metaphors in political discourse, particularly in showing how politicians employed dehumanizing frames when discussing immigrants (e.g., using speech that would generally frame ``animals,'' ``vermin,'' ``cargo,'' etc.)~\cite{CardChangEtAl2022,wright2025clifs}. As stated in the main paper, we leverage implicit metaphor detection to derive CLIFS' key measures.

Fusion Proximity ($f_{(I,T)}$) is calculated as the harmonic mean of the two directional identity scores ($S_{I \to T}$ and $S_{T \to I}$). This formulation ensures that the proximity score is maximized only when both directional signals are strong, thereby capturing the reciprocity of fusion~\cite{GomezBrooksEtAl2011,SwannKleinGomez2024,wright2025clifs}.  

\begin{equation}
    f_{(I,T)} \;=\;
    \frac{2 \, S_{I \to T} \, S_{T \to I}}
         {S_{I \to T} + S_{T \to I}}
\end{equation}

\noindent\ Directional proximity scores aim to capture the extent to which one conceptualizes the similarity of their fusion target to themselves, and vice versa. This is achieved by estimating the likelihood of fusion-target (e.g., social group) words, $T$, replacing identity (i.e., first-person singular pronouns; I, me, myself, etc.) words, $I$, and vice versa: identity words replacing fusion-target words. The set $T$ is a partially parameterized input and may be omitted. If no target list is provided, CLIFS still utilizes its base set of generic group terms; however, when the relevant groups or abstract targets are known, providing a custom list is strongly recommended. This user-specified list is automatically expanded to include semantically related terms. In the MTurk and field experiments, we used the following target set: \texttt{KNOWN\_GROUPS = ["religion", "religious", "church", "god", "faith"]}. In the violent manifesto experiment, we used an augmented target set based on that of the introductory paper, expanding it to include additional groups known to be present in the dataset: \texttt{KNOWN\_GROUPS = \{"religion", "religious", "church", "god", "college", "university", "school", "usa", "country", "america",\newline"white", "nation", "nazi", "culture"\}}~\cite{wright2025clifs}.

\begin{equation}
S_{x \to y} = 
  \frac{1}{M_y}
  \sum_{m=1}^{M_y}
    \sum_{w_v \in \mathcal{V}_x}
      P\bigl(w_v \mid C_m\bigr)^\alpha
\end{equation}

\noindent The directional scores, $S_{x \to y}$ ($x, y \in \{I, T\}$), represent the aggregate likelihood that words from category $x$ ($\mathcal{V}_x$) could replace words from category $y$ within a given context $C_m$. Concretely, all tokens in a text belonging to vocabulary $y$ are masked, yielding $M_y$ masked positions. For each masked token $m$, every candidate word $w_v \in \mathcal{V}_x$ is substituted in turn, and the probability of replacement is computed using the masked language model, $P\bigl(w_v \mid C_m\bigr)$ (i.e., the softmax distribution over the vocabulary at position $m$). Each probability is then raised to the smoothing parameter $\alpha$ to ensure numerical stability and enable meaningful aggregation~\cite{wright2025clifs}.

``Fictive-Kinship'' ($K_f$) is simply the directional score for kinship words replacing fusion-target words, $K_f = S_{K \to T}$.

CLIFS is tasked with predicting the VIFS score an individual would have received if they had completed the survey, using only their written text \cite{GomezBrooksEtAl2011, wright2025clifs}. This prediction can be framed either as a coarse-grained classification problem (low, medium, or high fusion) or as a fine-grained regression problem, producing a continuous score on the $1$--$7$ VIFS scale~\cite{GomezBrooksEtAl2011,wright2025clifs}. The model was trained and validated on a dataset of $N =984$ training samples and $N =246$ validation samples, and then evaluated on a held-out test set of $N =131$ samples. The dataset included 940 samples of real human data and 421 synthetic samples (generated through Round-Trip Translation (RTT) or Generative AI (GenAI)), spanning thirteen unique fusion-target categories\footnote{All $131$ test samples consisted exclusively of real human data}~\cite{AshokkumarPennebaker2022,FangXie2022,ZhangMiZhouEtAl2024,SwannKleinGomez2024,wright2025clifs}. In each case, participants (human or AI) were tasked with writing for 6--8 minutes about their relationship to a given fusion target. Their corresponding ground-truth identity fusion scores were obtained using the VIFS~\cite{GomezBrooksEtAl2011,AshokkumarPennebaker2022,wright2025clifs}.

CLIFS can be utilized on systems equipped with an NVIDIA GPU as well as on CPU-only systems. All experiments in this study were conducted using a GPU-enabled environment, specifically an NVIDIA RTX 4070 Ti (12 GB). Importantly, the classification model used for coarse-grained analysis employed fully local CLIFS implementations---specifically Random Forest variants rather than the Ensemble model, which integrates non-private industry APIs---ensuring that no data were transmitted externally and that all analyses preserved data security and privacy. The regression model used for fine-grained analysis is available only as a local implementation, and thus similarly poses no data-privacy concerns.

\subsection*{Manifesto Text Extraction}

Textual documents in Microsoft Word format were parsed using~\textsc{python-docx} \cite{python_docx}, while PDFs were processed with ~\textsc{PyPDF} to extract embedded text \cite{pypdf}. When PDFs contained no extractable text (e.g., scanned documents), pages were rasterized at 2$\times$ resolution using \textsc{PyMuPDF} \cite{pymupdf} and subjected to optical character recognition using the \textsc{Tesseract} engine \cite{smith2007overview} via the \textsc{pytesseract} Python interface \cite{pytesseract}. Several documents were already available as plain-text exports from DJVU sources and were ingested directly without further processing. MP4 files were transcoded to 16~kHz mono audio and transcribed using the \textsc{Vosk} offline speech recognition toolkit \cite{vosk} with the \texttt{vosk-model-en-us-0.42-gigaspeech} model, which is built on the \textsc{Kaldi} decoding framework \cite{povey2011kaldi} and employs acoustic and language models trained on the \textit{GigaSpeech} corpus \cite{chen2021gigaspeech}.

\subsection*{Statistical analysis and visualization.}
We compared CLIFS score distributions between \emph{Victim} and \emph{Ideologue} text segments using density-normalized histograms with KDEs~\cite{Waskom2021} and ECDFs~\cite{seabold2010statsmodels}. Central tendency and dispersion were summarized using medians and IQRs. Differences in means and medians (computed as Victim $-$ Ideologue) were quantified with nonparametric bootstrap percentile 95\% confidence intervals with 5{,}000 resamples~\cite{2020SciPy-NMeth}. We additionally report Cohen's $d$ and Cliff's $\delta$ (via the Mann--Whitney $U$ statistic) as effect size summaries, and the 1D Wasserstein distance~\cite{2020SciPy-NMeth} as a global measure of distributional dissimilarity. For component-level analyses, we restricted to segments labeled \emph{High Fusion} by the coarse-grained CLIFS classifier and visualized the four component distributions ($f_{(I,T)}$, $K_f$, $S_{T\rightarrow I}$, $S_{I\rightarrow T}$) with shared axis ranges within each component to facilitate direct comparison. Nonparametric methods were chosen due to non-normal (skewed and bimodal) distributions, heterogeneous sample sizes, and the text-derived nature of the measures.

\end{document}